%% file: main.tex
\documentclass[onefignum,onetabnum]{siamart171218}

\input{main_shared}
\usepackage{bm}
\usepackage{subfigure}
\usepackage{bold-extra}
\usepackage{booktabs}
\usepackage{mathtools}
\usepackage[frozencache,cachedir=.]{minted}
\floatname{algorithm}{Procedure}
\graphicspath{{./figs/}}

\ifpdf
\hypersetup{
  pdftitle={DeepXDE},
  pdfauthor={L. Lu, X. Meng, Z. Mao, and G. E. Karniadakis}
}
\fi

\begin{document}

\maketitle

\begin{abstract}
Deep learning has achieved remarkable success in diverse applications; however, its use in solving partial differential equations (PDEs) has emerged only recently. Here, we present an overview of physics-informed neural networks (PINNs), which embed a PDE into the loss of the neural network using automatic differentiation. The PINN algorithm is simple, and it can be applied to different types of PDEs, including integro-differential equations, fractional PDEs, and stochastic PDEs. Moreover, from the implementation point of view, PINNs solve inverse problems as easily as forward problems. We propose a new residual-based adaptive refinement (RAR) method to improve the training efficiency of PINNs. For pedagogical reasons, we compare the PINN algorithm to a standard finite element method. We also present a Python library for PINNs, DeepXDE, which is designed to serve both as an education tool to be used in the classroom as well as a research tool for solving problems in computational science and engineering. Specifically, DeepXDE can solve forward problems given initial and boundary conditions, as well as inverse problems given some extra measurements. DeepXDE supports complex-geometry domains based on the technique of constructive solid geometry, and enables the user code to be compact, resembling closely the mathematical formulation. We introduce the usage of DeepXDE and its customizability, and we also demonstrate the capability of PINNs and the user-friendliness of DeepXDE for five different examples. More broadly, DeepXDE contributes to the more rapid development of the emerging Scientific Machine Learning field.
\end{abstract}



\section{Introduction}

In the last 15 years, deep learning in the form of deep neural networks (NNs), has been used very effectively in diverse applications~\cite{lecun2015deep}, such as computer vision and natural language processing. Despite the remarkable success in these and related areas, deep learning has not yet been widely used in the field of scientific computing. However, more recently, solving partial differential equations (PDEs), e.g., in the standard differential form or in the integral form, via deep learning has emerged
as a potentially new sub-field under the name of Scientific Machine Learning (SciML)~\cite{baker2019workshop}. In particular, we can replace traditional numerical discretization methods with a neural network that approximates the solution to a PDE.

To obtain the approximate solution of a PDE via deep learning, a key step is to constrain the neural network to minimize the PDE residual, and several approaches have been proposed to accomplish this. Compared to the traditional mesh-based methods, such as the finite difference method (FDM) and the finite element method (FEM), deep learning could be a mesh-free approach by taking advantage of the automatic differentiation~\cite{raissi2019physics}, and could break the curse of dimensionality~\cite{poggio2017and,grohs2018proof}. Among these approaches, some can only be applied to particular types of problems, such as image-like input domain~\cite{khoo2017solving,long2018pde,zhu2019physics} or parabolic PDEs~\cite{beck2017machine,han2018solving}. Some researchers adopt the variational form of PDEs and minimize the corresponding energy functional~\cite{weinan2018deep,he2018relu}. However, not all PDEs can be derived from a known functional, and thus Galerkin type projections have also been considered~\cite{meade1994numerical}. Alternatively, one could use the PDE in strong form directly~\cite{dissanayake1994neural,van1995neural,lagaris1998artificial,lagaris2000neural,berg2018unified,sirignano2018dgm,raissi2019physics}; in this form, automatic differentiation could be used directly to avoid truncation errors and the numerical quadrature errors of variational forms. This strong form approach was introduced in \cite{raissi2019physics} coining the term physics-informed neural networks (PINNs). An attractive feature of PINNs is that it can be used to solve inverse problems with minimum change of the code for forward problems~\cite{raissi2019physics,raissi2020hidden,tartakovsky2018learning,he2019physics,chen2019physics}. In addition, PINNs have been further extended to solve integro-differential equations (IDEs), fractional differential equations (FDEs)~\cite{pang2019fpinns}, and stochastic differential equations (SDEs)~\cite{zhang2019quantifying,yang2018physics,nabian2018deep,zhang2019learning}.

In this paper, we present various PINN algorithms implemented in a Python library DeepXDE\footnote{Source code is published under the Apache License, Version 2.0 on GitHub. \url{https://github.com/lululxvi/deepxde}}, which is designed to serve both as an education tool to be used in the classroom as well as a research tool for solving problems in computational science and engineering (CSE). DeepXDE can be used to solve multi-physics problems, and supports complex-geometry domains based on the technique of constructive solid geometry (CSG), hence avoiding tedious and time-consuming computational geometry tasks. By using DeepXDE, time-dependent PDEs can be solved as easily as steady states by only defining the initial conditions. In addition to the main workflow of DeepXDE, users can readily monitor and modify the solution process via \texttt{callback} functions, e.g., monitoring the Fourier spectrum of the neural network solution, which can reveal the learning mode of the NN \cref{fig:frequency}. Last but not least, DeepXDE is designed to make the user code stay compact and manageable, resembling closely the mathematical formulation.

The paper is organized as follows. In \cref{sec:pinn}, after briefly introducing deep neural networks and automatic differentiation, we present the algorithm, approximation theory, and error analysis of PINNs, and make a comparison between PINNs and FEM. We then discuss how to use PINNs to solve integro-differential equations and inverse problems. In addition, we propose the residual-based adaptive refinement (RAR) method to improve the training efficiency of PINNs. In \cref{sec:deepxde}, we introduce the usage of our library, DeepXDE, and its customizability. In \cref{sec:example}, we demonstrate the capability of PINNs and friendly use of DeepXDE for five different examples. Finally, we conclude the paper in \cref{sec:conc}.

\section{Algorithm and theory of physics-informed neural networks}
\label{sec:pinn}

In this section, we first provide a brief overview of deep neural networks and automatic differentiation, and present the algorithm and theory of PINNs for solving PDEs. We then make a comparison between PINNs and FEM, and discuss how to use PINNs to solve integro-differential equations and inverse problems. Next we propose RAR, an efficient way to select the residual points adaptively during the training process.

\subsection{Deep neural networks}

Mathematically, a deep neural network is a particular choice of a compositional function. The simplest neural network is the feed-forward neural network (FNN), also called multilayer perceptron (MLP), which applies linear and nonlinear transformations to the inputs recursively. Although many different types of neural networks have been developed in the past decades, such as the convolutional neural network and the recurrent neural network. In this paper we consider FNN, which is sufficient for most PDE problems, and residual neural network (ResNet), which is easier to train for deep networks. However, it is straightforward to employ other types of neural networks.

Let $\mathcal{N}^L(\mathbf{x}): \mathbb{R}^{d_{\text{in}}} \to \mathbb{R}^{d_{\text{out}}}$ be a $L$-layer neural network, or a $(L-1)$-hidden layer neural network, with $N_\ell$ neurons in the $\ell$-th layer ($N_0 = d_{\text{in}}$, $N_L = d_{\text{out}}$). Let us denote the weight matrix and bias vector in the $\ell$-th layer by $\bm{W}^\ell \in \mathbb{R}^{N_\ell \times N_{\ell-1}}$ and  $\mathbf{b}^\ell \in \mathbb{R}^{N_\ell}$, respectively. Given a nonlinear activation function $\sigma$,  which is applied element-wisely, the FNN is recursively defined as follows:
\begin{align*}
    \text{input layer:} & \quad \mathcal{N}^0(\textbf{x}) = \textbf{x} \in \mathbb{R}^{d_{\text{in}}}, \\
    \text{hidden layers:} & \quad \mathcal{N}^\ell(\textbf{x}) = \sigma(\bm{W}^{\ell}\mathcal{N}^{\ell-1}(\textbf{x}) + \bm{b}^{\ell}) \in \mathbb{R}^{N_\ell}, \quad \text{for} \quad 1 \le \ell \le L-1, \\
    \text{output layer:} & \quad \mathcal{N}^{L}(\textbf{x}) = \bm{W}^{L}\mathcal{N}^{L-1}(\textbf{x}) + \bm{b}^{L} \in \mathbb{R}^{d_{\text{out}}};
\end{align*}
see also a visualization of a neural network in \cref{fig:pinn}. Commonly used activation functions include the logistic sigmoid $1/(1+e^{-x})$, the hyperbolic tangent ($\tanh$), and the rectified linear unit (ReLU, $\max\{x, 0\}$).

\subsection{Automatic differentiation}
In PINNs, it is required to compute the derivatives of the network outputs with respect to the network inputs. There are four possible methods for computing the derivatives \cite{baydin2017automatic,margossian2019review}: (1) hand-coded analytical derivative; (2) finite difference or other numerical approximations; (3) symbolic differentiation (used in software programs such as Mathematica, Maxima, and Maple); and (4) automatic differentiation (AD, also called algorithmic differentiation). In deep learning, the derivatives are evaluated using backpropagation \cite{rumelhart1986learning}, a specialized technique of AD.

Considering the fact that the neural network represents a compositional function, then AD applies the chain rule repeatedly to compute the derivatives. There are two steps in AD: one forward pass to compute the values of all variables, and one backward pass to compute the derivatives. To demonstrate AD, we consider a FNN of only one hidden layer with two inputs $x_1$ and $x_2$ and one output $y$:
\begin{align*}
    v &= -2 x_1 + 3 x_2 + 0.5, \\
    h &= \tanh v, \\
    y &= 2 h - 1.
\end{align*}
The forward pass and backward pass of AD for computing the partial derivatives $\frac{\partial y}{\partial x_1}$ and $\frac{\partial y}{\partial x_2}$ at $(x_1, x_2) = (2,1)$ are shown in Table \ref{tab:ad}.

We can see that AD only requires one forward pass and one backward pass to compute all the partial derivatives, no matter what the input dimension is. In contrast, using finite differences computing each partial derivative $\frac{\partial y}{\partial x_i}$ requires two function valuations $y(x_1, \dots, x_i, \dots, x_{d_{\text{in}}})$ and $y(x_1, \dots, x_i + \Delta x_i, \dots, x_{d_{\text{in}}})$ for some small number $\Delta x_i$, and thus in total $d_{\text{in}}+1$ forward passes are required to evaluate all the partial derivatives. Hence, AD is much more efficient than finite difference when the input dimension is high (see \cite{baydin2017automatic,margossian2019review} for more details of the comparison between AD and the other three methods). To compute $n^{th}$-order derivatives, AD can be applied recursively $n$ times. However, this nested approach may lead to inefficiency and numerical instability, and hence other methods, e.g., Taylor-Mode AD, have been developed for this purpose \cite{betancourt2018geometric,bettencourt2019taylor}. Finally we note that with AD we differentiate the NN and therefore we can deal with noisy data \cite{pang2019fpinns}.

\begin{table}[htbp]
{\footnotesize
  \caption{Example of AD to compute the partial derivatives $\frac{\partial y}{\partial x_1}$ and $\frac{\partial y}{\partial x_2}$ at $(x_1, x_2) = (2,1)$.} \label{tab:ad}
\begin{center}
  \begin{tabular}{l|l} \hline
  Forward pass & Backward pass \\ \hline
  $x_1 = 2$ & $\frac{\partial y}{\partial y} = 1$ \\
  $x_2 = 1$ &  \\
  \hline
  $v = -2 x_1 + 3 x_2 + 0.5 = -0.5$ & $\frac{\partial y}{\partial h} = \frac{\partial (2h-1)}{\partial h} = 2$ \\
  $h = \tanh v \approx -0.462$ & $\frac{\partial y}{\partial v} = \frac{\partial y}{\partial h} \frac{\partial h}{\partial v} = \frac{\partial y}{\partial h} \text{sech}^2 (v) \approx 1.573$ \\
  \hline
  $y = 2 h -1 = -1.924$ & $\frac{\partial y}{\partial x_1} = \frac{\partial y}{\partial v}\frac{\partial v}{\partial x_1} = \frac{\partial y}{\partial v} \times (-2) = -3.146$ \\
  & $\frac{\partial y}{\partial x_2} = \frac{\partial y}{\partial v}\frac{\partial v}{\partial x_2} = \frac{\partial y}{\partial v} \times 3 = 4.719$ \\
  \hline
  \end{tabular}
\end{center}
}
\end{table}

\subsection{Physics-informed neural networks (PINNs) for solving PDEs}\label{sec:pinnalgo}
We consider the following PDE parameterized by $\bm{\lambda}$ for the solution $u(\mathbf{x})$ with $\mathbf{x} = (x_1, \dots, x_d)$ defined on a domain $\Omega \subset \mathbb{R}^d$:
\begin{equation}\label{eq:pde}
f\left( \mathbf{x}; \frac{\partial u}{\partial x_1},  \dots, \frac{\partial u}{\partial x_d}; \frac{\partial^2 u}{\partial x_1\partial x_1},  \dots, \frac{\partial^2 u}{\partial x_1\partial x_d};\dots; \bm{\lambda} \right) = 0, \quad \mathbf{x} \in \Omega,
\end{equation}
with suitable boundary conditions
\begin{equation*}
\mathcal{B}(u, \mathbf{x}) = 0 \quad \text{on} \quad \partial \Omega,
\end{equation*}
where $\mathcal{B}(u, \mathbf{x})$ could be Dirichlet, Neumann, Robin, or periodic boundary conditions. For time-dependent problems, we consider time $t$ as a special component of $\mathbf{x}$, and $\Omega$ contains the temporal domain. The initial condition can be simply treated as a special type of Dirichlet boundary condition on the spatio-temporal domain.

\begin{figure}[htbp]
    \centering
    \includegraphics{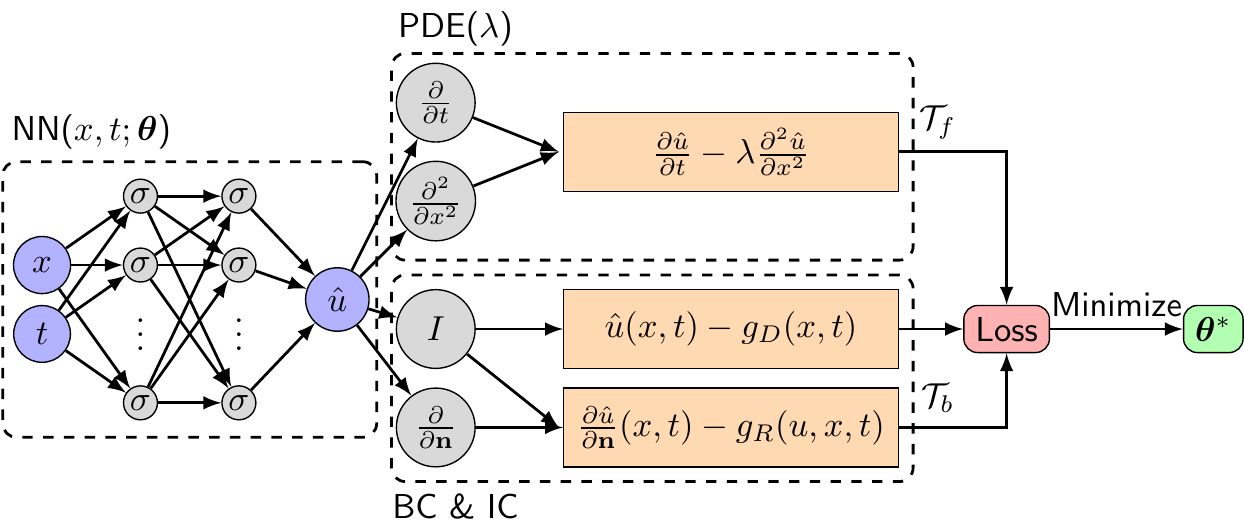}
    \caption{Schematic of a PINN for solving the diffusion equation $\frac{\partial u}{\partial t} = \lambda \frac{\partial^2 u}{\partial x^2}$ with mixed boundary conditions (BC) $u(x,t) = g_D(x,t)$ on $\Gamma_D \subset \partial \Omega$ and $\frac{\partial u}{\partial \mathbf{n}}(x,t) = g_R(u, x,t)$ on $\Gamma_R \subset \partial \Omega$. The initial condition (IC) is treated as a special type of boundary conditions. $\mathcal{T}_f$ and $\mathcal{T}_b$ denote the two sets of residual points for the equation and BC/IC.}
    \label{fig:pinn}
\end{figure}

\begin{algorithm}[htbp]
\caption{The PINN algorithm for solving differential equations.}\label{alg:PINN}
\begin{enumerate}
\item[Step 1] Construct a neural network $\hat{u}(\mathbf{x};\bm{\theta})$ with parameters $\bm{\theta}$.
\item[Step 2] Specify the two training sets $\mathcal{T}_f$ and $\mathcal{T}_b$ for the equation and boundary/initial conditions.
\item[Step 3] Specify a loss function by summing the weighted $L^2$ norm of both the PDE equation and boundary condition residuals.
\item[Step 4] Train the neural network to find the best parameters $\bm{\theta}^*$ by minimizing the loss function $\mathcal{L}(\bm{\theta};\mathcal{T}).$
\end{enumerate}
\end{algorithm}

The algorithm of PINN~\cite{lagaris2000neural,raissi2019physics} is shown in Procedure~\ref{alg:PINN}, and visually in the schematic of \cref{fig:pinn} solving a diffusion equation $\frac{\partial u}{\partial t} = \lambda \frac{\partial^2 u}{\partial x^2}$ with mixed boundary conditions $u(x,t) = g_D(x,t)$ on $\Gamma_D \subset \partial \Omega$ and $\frac{\partial u}{\partial \mathbf{n}}(x,t) = g_R(u, x,t)$ on $\Gamma_R \subset \partial \Omega$. We explain each step as follows. In a PINN, we first construct a neural network $\hat{u}(\mathbf{x}; \bm{\theta})$ as a surrogate of the solution $u(\mathbf{x})$, which takes the input $\mathbf{x}$ and outputs a vector with the same dimension as $u$. Here, $\bm{\theta} = \{\bm{W}^\ell, \bm{b}^\ell\}_{1 \le \ell \le L}$ is the set of all weight matrices and bias vectors in the neural network $\hat{u}$. One advantage of PINNs by choosing neural networks as the surrogate of $u$ is that we can take the derivatives of $\hat{u}$ with respect to its input $\mathbf{x}$ by applying the chain rule for differentiating compositions of functions using the automatic differentiation (AD), which is conveniently integrated in machine learning packages, such as TensorFlow~\cite{abadi2016tensorflow} and PyTorch~\cite{paszke2017automatic}.

In the next step, we need to restrict the neural network $\hat{u}$ to satisfy the physics imposed by the PDE and boundary conditions. In practice, we restrict $\hat{u}$ on some scattered points (e.g., randomly distributed points, or clustered points in the domain \cite{MJK2020CMAME}), i.e., the training data $\mathcal{T} = \{\mathbf{x}_1, \mathbf{x}_2, \dots, \mathbf{x}_{|\mathcal{T}|}\}$ of size $|\mathcal{T}|$. In addition, $\mathcal{T}$ is comprised of two sets $\mathcal{T}_f \subset \Omega$ and $\mathcal{T}_b \subset \partial \Omega$, which are the points in the domain and on the boundary, respectively. We refer $\mathcal{T}_f$ and $\mathcal{T}_b$ as the sets of ``residual points''.

To measure the discrepancy between the neural network $\hat{u}$ and the constraints, we consider the loss function defined as the weighted summation of the $L^2$ norm of residuals for the equation and boundary conditions:
\begin{equation}\label{eq:loss}
\mathcal{L}(\bm{\theta}; \mathcal{T}) = w_f \mathcal{L}_f(\bm{\theta}; \mathcal{T}_f) + w_b \mathcal{L}_{b}(\bm{\theta}; \mathcal{T}_b),
\end{equation}
where
\begin{align*}
\mathcal{L}_f(\bm{\theta}; \mathcal{T}_f) &= \frac{1}{|\mathcal{T}_f|} \sum_{\mathbf{x} \in \mathcal{T}_f} \left\Vert f\left( \mathbf{x}; \frac{\partial \hat{u}}{\partial x_1},  \dots, \frac{\partial \hat{u}}{\partial x_d}; \frac{\partial^2 \hat{u}}{\partial x_1\partial x_1},  \dots, \frac{\partial^2 \hat{u}}{\partial x_1\partial x_d};\dots;\bm{\lambda} \right) \right\Vert_2^2, \\
\mathcal{L}_{b}(\bm{\theta}; \mathcal{T}_b) &= \frac{1}{|\mathcal{T}_b|} \sum_{\mathbf{x} \in \mathcal{T}_b} \| \mathcal{B}(\hat{u}, \mathbf{x}) \|_2^2,
\end{align*}
and $w_f$ and $w_b$ are the weights.
The loss involves derivatives, such as the partial derivative $\partial \hat{u}/\partial x_1$ or the normal derivative at the boundary $\partial \hat{u}/\partial \mathbf{n} = \nabla \hat{u} \cdot \mathbf{n}$, which are handled via AD.

In the last step, the procedure of searching for a good $\bm{\theta}$ by minimizing the loss $\mathcal{L}(\bm{\theta}; \mathcal{T})$ is called ``training''. Considering the fact that the loss is highly nonlinear and non-convex with respect to $\bm{\theta}$ \cite{blum1989training}, we usually minimize the loss function by gradient-based optimizers, such as gradient descent, Adam~\cite{kingma2014adam}, and L-BFGS~\cite{byrd1995limited}. We remark that based on our experience, for smooth PDE solutions L-BFGS can find a good solution with less iterations than Adam, because L-BFGS uses second-order derivatives of the loss function, while Adam only relies on first-order derivatives. However, for stiff solutions L-BFGS is more likely to be stuck at a bad local minimum. The required number of iterations highly depends on the problem (e.g., the smoothness of the solution), and to check whether the network converges or not, we can monitor the loss function or the PDE residual using \texttt{callback} functions. We also note that acceleration of training can be achieved by using adaptive activation function that may remove bad local minima, see \cite{jagtap2020adaptive,jagtap2019locally}.

Unlike traditional numerical methods, for PINNs there is no guarantee of unique solutions, because PINN solutions are obtained by solving non-convex optimization problems, which in general do not have a unique solution. In practice, to achieve a good level of accuracy, we need to tune all the hyperparameters, e.g., network size, learning rate, and the number of residual points. The required network size depends highly on the smoothness of the PDE solution. For example, a small network (e.g., a few layers and twenty neurons per layer) is sufficient for solving the 1D Poisson equation, but a deeper and wider network is required for the 1D Burgers equation to achieve a similar level of accuracy. We also note that PINNs may converge to different solutions from different network initial values \cite{pang2019fpinns,tartakovsky2018learning,he2019physics}, and thus a common strategy is that we train PINNs from random initialization for a few times (e.g., 10 independent runs) and choose the network with the smallest training loss as the final solution.

In the algorithm of PINN introduced above, we enforce soft constraints of boundary/initial conditions through the loss $\mathcal{L}_b$. This approach can be used for complex domains and any type of boundary conditions. On the other hand, it is possible to enforce hard constraints for simple cases \cite{lagaris1998artificial}. For example, when the boundary condition is $u(0)=u(1)=0$ with $\Omega = [0, 1]$, we can simply choose the surrogate model as $\hat{u}(x) = x(x-1)\mathcal{N}(x)$ to satisfy the boundary condition automatically, where $\mathcal{N}(x)$ is a neural network.

We note that we have great flexibility in choosing the residual points $\mathcal{T}$, and here we provide three possible strategies:
\begin{enumerate}
    \item We can specify the residual points at the beginning of training, which could be grid points on a lattice or random points, and never change them during the training process.
    \item In each optimization iteration, we could select randomly different residual points.
    \item We could improve the location of the residual points adaptively during training, e.g., the method proposed in \cref{sec:rar}.
\end{enumerate}
When the number of residual points required is very large, e.g., in multiscale problems, it is computationally expensive to calculate the loss and gradient in every iteration. Instead of using all residual points, we can split the residual points into small batches, and in each iteration we only use one batch to calculate the loss and update model parameters; this is the so-called ``mini-batch'' gradient descent. The aforementioned strategy (2), i.e., re-sampling in each step, is a special case of mini-batch gradient descent by choosing $\mathcal{T} = \Omega$ with $|\mathcal{T}| = \infty$.

\begin{figure}[htbp]
    \centering
    \includegraphics[width=0.9\textwidth]{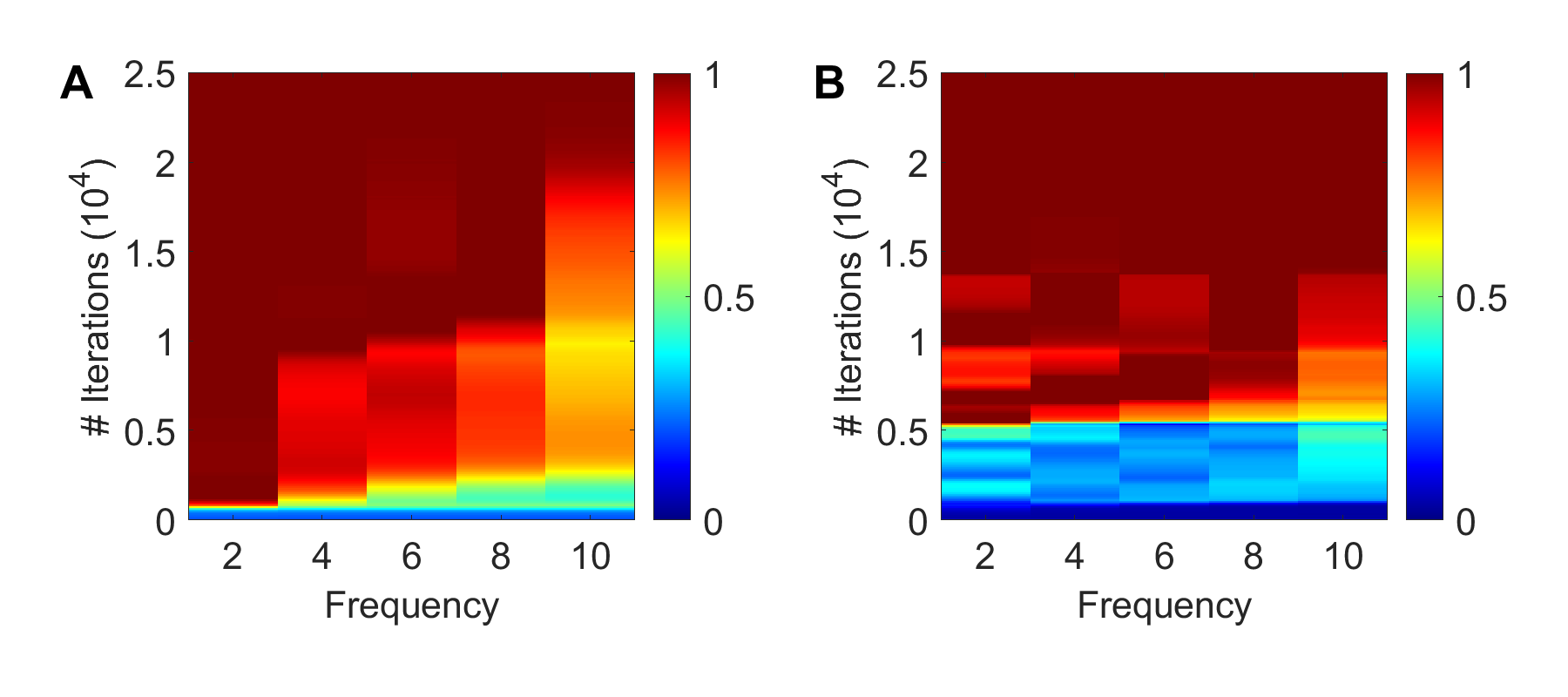}
    \caption{Convergence of the amplitude for each frequency during the training process. (\textbf{A}) A neural network is trained to approximate the function $f(x) = \sum_{k=1}^5 \sin(2kx)/(2k)$. The color represents amplitude values with the maximum amplitude for each frequency normalized to 1. (\textbf{B}) A PINN is used to solve the Poisson equation $- f_{xx} = \sum_{k=1}^5 2k \sin(2kx)$ with zero boundary conditions. We use a neural network of 4 hidden layers and 20 neurons per layer. The learning rate is chosen as $10^{-4}$, and 500 random points are sampled for training.}
    \label{fig:frequency}
\end{figure}

Recent studies show that for function approximation, neural networks learn target functions from low to high frequencies \cite{rahaman2018spectral,xu2019frequency}, but here we show that the learning mode of PINNs is different due to the existence of high-order derivatives. For example, when we approximate the function $f(x) = \sum_{k=1}^5 \sin(2kx)/(2k)$ in $[-\pi ,\pi]$ by a NN, the function is learned from low to high frequency (\cref{fig:frequency}A). However, when we employ a PINN to solve the Poisson equation $- f_{xx} = \sum_{k=1}^5 2k \sin(2kx)$ with zero boundary conditions in the same domain, all frequencies are learned almost simultaneously (\cref{fig:frequency}B). Interestingly, by comparing \cref{fig:frequency}A and \cref{fig:frequency}B we can see that at least in this case solving the PDE using a PINN is faster than approximating a function using a NN. We can monitor this training process using the \texttt{callback} functions in our library DeepXDE as discussed later.

\subsection{Approximation theory and error analysis for PINNs}
\label{sec:theory}

One fundamental question related to PINNs is whether there exists a neural network satisfying both the PDE equation and the boundary conditions, i.e., whether there exists a neural network that can simultaneously and uniformly approximate a function and its partial derivatives. To address this question, we first introduce some notation. Let $\mathbb{Z}_+^d$ be the set of $d$-dimensional nonnegative integers. For $\mathbf{m} = (m_1, \dots, m_d) \in \mathbb{Z}_+^d$, we set $|\mathbf{m}| \coloneqq m_1+\dots +m_d$, and
$$D^{\mathbf{m}} \coloneqq \frac{\partial^{|\mathbf{m}|}}{\partial x_1^{m_1}\dots \partial x_d^{m_d}}.$$
We say $f \in C^{\mathbf{m}}(\mathbb{R}^d)$ if $D^\mathbf{k}f\in C(\mathbb{R}^d)$ for all $\mathbf{k} \leq \mathbf{m}$, $\mathbf{k} \in \mathbb{Z}_+^d$, where $C(\mathbb{R}^d) = \{ f: \mathbb{R}^d \to \mathbb{R} | f \text{ is continuous} \}$ is the space of continuous functions. Then, we recall the following theorem of derivative approximation using single hidden layer neural networks due to Pinkus \cite{pinkus1999approximation}.

\begin{theorem}\label{thm:app}
Let $\mathbf{m}^i \in \mathbb{Z}_+^d$, $i=1,\dots, s$, and set $m=\max_{i=1,\dots,s} |\mathbf{m}^i|$. Assume $\sigma \in C^m(\mathbb{R})$ and $\sigma$ is not a polynomial. Then the space of single hidden layer neural nets
$$\mathcal{M}(\sigma) \coloneqq span\{ \sigma(\mathbf{w}\cdot \mathbf{x} + b): \mathbf{w}\in \mathbb{R}^d,b\in\mathbb{R} \}$$
is dense in
$$C^{\mathbf{m}^1,\dots,\mathbf{m}^s}(\mathbb{R}^d) \coloneqq \cap_{i=1}^sC^{\mathbf{m}^i}(\mathbb{R}^d),$$
i.e., for any $f \in C^{\mathbf{m}^1,\dots,\mathbf{m}^s}(\mathbb{R}^d)$, any compact $K \subset \mathbb{R}^d$, and any $\varepsilon > 0$, there exists a $g \in \mathcal{M}(\sigma)$ satisfying
$$\max_{\mathbf{x} \in K} | D^\mathbf{k}f(\mathbf{x})-D^\mathbf{k}g(\mathbf{x}) | < \varepsilon,$$
for all $\mathbf{k} \in \mathbb{Z}_+^d$ for which $\mathbf{k}\leq \mathbf{m}^i$ for some $i$.
\end{theorem}

\begin{figure}
    \centering
    \includegraphics{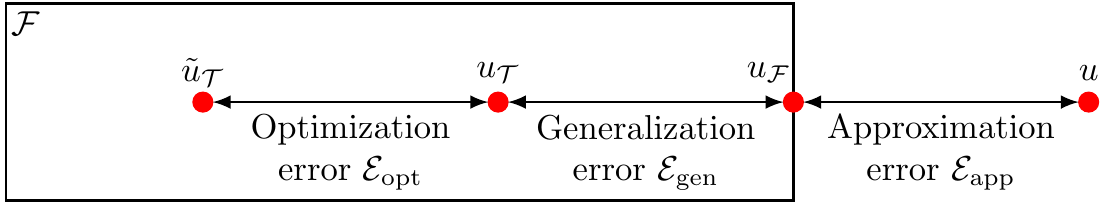}
    \caption{Illustration of errors of a PINN. The total error consists of the approximation error, the optimization error, and the generalization error. Here, $u$ is the PDE solution, $u_\mathcal{F}$ is the best function close to $u$ in the function space $\mathcal{F}$, $u_\mathcal{T}$ is the neural network whose loss is at a global minimum, and $\tilde{u}_{\mathcal{T}}$ is the function obtained by training a neural network.}
    \label{fig:error}
\end{figure}

\Cref{thm:app} shows that feed-forward neural nets with enough neurons can simultaneously and uniformly approximate any function and its partial derivatives. However, neural networks in practice have limited size. Let $\mathcal{F}$ denote the family of all the functions that can be represented by our chosen neural network architecture. The solution $u$ is unlikely to belong to the family $\mathcal{F}$, and we define $u_{\mathcal{F}} = \arg\min_{f \in \mathcal{F}} \| f - u\|$ as the best function in $\mathcal{F}$ close to $u$ (\cref{fig:error}). Because we only train the neural network on the training set $\mathcal{T}$, we define $u_{\mathcal{T}} = \arg\min_{f \in \mathcal{F}} \mathcal{L}(f;\mathcal{T})$ as the neural network whose loss is at global minimum. For simplicity, we assume that $u$, $u_\mathcal{F}$ and $u_{\mathcal{T}}$ are well defined and unique. Finding $u_\mathcal{T}$ by minimizing the loss is often computationally intractable \cite{blum1989training}, and our optimizer returns an approximate solution $\tilde{u}_{\mathcal{T}}$.

We can then decompose the total error $\mathcal{E}$ as~\cite{bottou2008tradeoffs}
\begin{equation*}
    \mathcal{E} \coloneqq \|\tilde{u}_\mathcal{T} - u\| \leq 
    \underbrace{\|\tilde{u}_\mathcal{T} - u_\mathcal{T}\|}_{\mathcal{E}_{\text{opt}}} +
    \underbrace{\|u_\mathcal{T} - u_\mathcal{F}\|}_{\mathcal{E}_{\text{gen}}} +
    \underbrace{\|u_\mathcal{F} - u\|}_{\mathcal{E}_{\text{app}}}.
\end{equation*}
The approximation error $\mathcal{E}_{\text{app}}$ measures how closely $u_\mathcal{F}$ can approximate $u$. The generalization error $\mathcal{E}_{\text{gen}}$ is determined by the number/locations of residual points in $\mathcal{T}$ and the capacity of the family $\mathcal{F}$. Neural networks of larger size have smaller approximation errors but could lead to higher generalization errors, which is called bias-variance tradeoff. Overfitting occurs when the generalization error dominates. In addition, the optimization error $\mathcal{E}_{\text{opt}}$ stems from the loss function complexity and the optimization setup, such as learning rate and number of iterations. However, currently there is no error estimation for PINNs yet, and even quantifying the three errors for supervised learning is still an open research problem \cite{lu2018collapse,lu2019dying,jin2019quantifying}.

\subsection{Comparison between PINNs and FEM}

To further explain the ideas of PINNs and to help those with the knowledge of FEM understand PINNs more easily, we make a comparison between PINNs and FEM point by point (\cref{tab:pinn_fem}):
\begin{itemize}
    \item In FEM we approximate the solution $u$ by a piecewise polynomial with point values to be determined, while in PINNs we construct a neural network as the surrogate model parameterized by weights and biases.
    \item FEM typically requires a mesh generation, while PINN is totally mesh-free, and we can use either a grid or random points.
    \item FEM converts a PDE to an algebraic system, using the stiffness matrix and mass matrix, while PINN embeds the PDE and boundary conditions into the loss function.
    \item In the last step, the algebraic system in FEM is solved exactly by a linear solver, but the network in PINN is learned by a gradient-based optimizer.
\end{itemize}
At a more fundamental level, PINNs provide a nonlinear approximation to the function and its derivatives whereas FEM represent a linear approximation.

\begin{table}[htbp]
{\footnotesize
  \caption{Comparison between PINN and FEM.} \label{tab:pinn_fem}
\begin{center}
  \begin{tabular}{c|cc} \hline
   & PINN & FEM \\ \hline
    Basis function & Neural network (nonlinear) & Piecewise polynomial (linear) \\
    Parameters & Weights and biases & Point values \\
    Training points & Scattered points (mesh-free) & Mesh points \\
    PDE embedding & Loss function & Algebraic system \\
    Parameter solver & Gradient-based optimizer & Linear solver \\
    Errors & $\mathcal{E}_{\text{app}}$, $\mathcal{E}_{\text{gen}}$ and $\mathcal{E}_{\text{opt}}$ (\cref{sec:theory}) & Approximation/quadrature errors \\
    Error bounds & Not available yet & Partially available \cite{ciarlet2002finite,johnson2012numerical} \\
    \hline
  \end{tabular}
\end{center}
}
\end{table}

\subsection{PINNs for solving integro-differential equations}
\label{sec:ide}
\begin{figure}[htbp]
    \centering
    \includegraphics{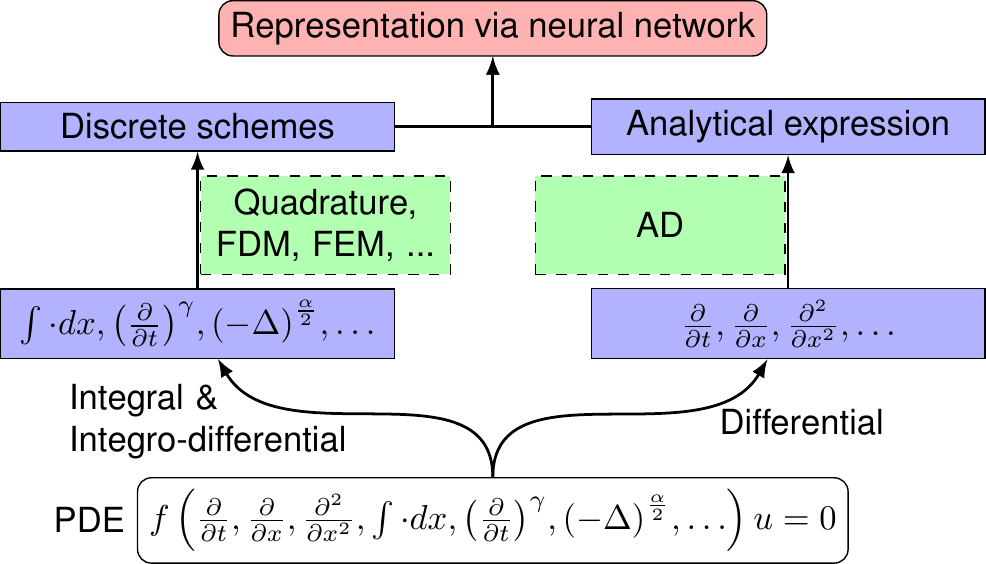}
    \caption{Schematic illustrating the modificaiton of the PINN algorithm for solving integro-differential equations. We employ the automatic differentiation technique to analytically derive the integer-order derivatives, and we approximate integral operators numerically using standard methods. (The figure is reproduced from \cite{pang2019fpinns}.)}
    \label{fig:fpinn}
\end{figure}

When solving integro-differential equations (IDEs), we still employ the automatic differentiation technique to analytically derive the integer-order derivatives, while we approximate integral operators numerically using classical methods (\cref{fig:fpinn})~\cite{pang2019fpinns}, such as Gaussian quadrature. Therefore, we introduce a fourth error component, the discretization error $\mathcal{E}_{\text{dis}}$, due to the approximation of the integral by Gaussian quadrature.

For example, when solving
$$\frac{dy}{dx} + y(x) = \int_0^x e^{t-x}y(t)dt,$$
we first use Gaussian quadrature of degree $n$ to approximate the integral
$$\int_0^x e^{t-x}y(t)dt \approx \sum_{i=1}^n w_i e^{t_i(x)-x}y(t_i(x)),$$
and then we use a PINN to solve the following PDE instead of the original equation
$$\frac{dy}{dx} + y(x) \approx \sum_{i=1}^n w_i e^{t_i(x)-x}y(t_i(x)).$$
PINNs can also be easily extended to solve FDEs~\cite{pang2019fpinns} and SDEs~\cite{zhang2019quantifying,yang2018physics,nabian2018deep,zhang2019learning}, but we do not discuss here 
such cases due to the page limit.

\subsection{PINNs for solving inverse problems}

In inverse problems, there are some unknown parameters $\bm{\lambda}$ in \cref{eq:pde}, but we have some extra information on some points $\mathcal{T}_i \subset \Omega$ besides the differential equation and boundary conditions:
$$\mathcal{I}(u, \mathbf{x}) = 0, \quad \text{for} \quad \mathbf{x} \in \mathcal{T}_i.$$
From the implementation point of view, PINNs solve inverse problems as easily as forward problems \cite{raissi2019physics,pang2019fpinns}. The only difference between solving forward and inverse problems is that we add an extra loss term to \cref{eq:loss}:
\begin{equation*}
\mathcal{L}(\bm{\theta}, \bm{\lambda}; \mathcal{T}) = w_f \mathcal{L}_f(\bm{\theta},\bm{\lambda}; \mathcal{T}_f) + w_b \mathcal{L}_{b}(\bm{\theta},\bm{\lambda}; \mathcal{T}_b) + w_i \mathcal{L}_{i}(\bm{\theta},\bm{\lambda}; \mathcal{T}_i),
\end{equation*}
where
\begin{equation*}
\mathcal{L}_{i}(\bm{\theta},\bm{\lambda}; \mathcal{T}_i) = \frac{1}{|\mathcal{T}_i|} \sum_{\mathbf{x} \in \mathcal{T}_i} \| \mathcal{I}(\hat{u}, \mathbf{x}) \|_2^2.
\end{equation*}
We then optimize $\bm{\theta}$ and $\bm{\lambda}$ together, and our solution is $\bm{\theta}^*, \bm{\lambda}^*=\arg\min_{\bm{\theta},\bm{\lambda}} \mathcal{L}(\bm{\theta},\bm{\lambda};\mathcal{T})$.

\subsection{Residual-based adaptive refinement (RAR)}
\label{sec:rar}

As we discussed in \cref{sec:pinnalgo}, the residual points $\mathcal{T}$ are usually randomly distributed in the domain. This works well for most cases, but it may not be efficient for certain PDEs that exhibit solutions with steep gradients. Take the Burgers equation as an example, intuitively we should put more points near the sharp front to capture the discontinuity well. However, it is challenging, in general, to design a good distribution of residual points for problems whose solution is unknown. To overcome this challenge, we propose a residual-based adaptive refinement (RAR) method to improve the distribution of residual points during training process (Procedure~\ref{alg:rat}), conceptually similar to FEM refinement methods \cite{ainsworth2011posteriori}. The idea of RAR is that we will add more residual points in the locations where the PDE residual $\left\Vert f\left( \mathbf{x}; \frac{\partial \hat{u}}{\partial x_1},  \dots, \frac{\partial \hat{u}}{\partial x_d}; \frac{\partial^2 \hat{u}}{\partial x_1\partial x_1},  \dots, \frac{\partial^2 \hat{u}}{\partial x_1\partial x_d};\dots;\bm{\lambda} \right) \right\Vert$ is large, and we repeat adding points until the mean residual
\begin{equation}\label{eq:rat}
    \mathcal{E}_r = \frac{1}{V} \int_{\Omega} \left\Vert f\left( \mathbf{x}; \frac{\partial \hat{u}}{\partial x_1},  \dots, \frac{\partial \hat{u}}{\partial x_d}; \frac{\partial^2 \hat{u}}{\partial x_1\partial x_1}, \dots, \frac{\partial^2 \hat{u}}{\partial x_1\partial x_d};\dots;\bm{\lambda} \right) \right\Vert d\mathbf{x}
\end{equation}
is smaller than a threshold $\mathcal{E}_0$, where $V$ is the volume of $\Omega$.

\begin{algorithm}[htbp]
\caption{RAR for improving the distribution of residual points for training.}\label{alg:rat}
\begin{enumerate}
\item[Step 1] Select the initial residual points $\mathcal{T}$, and train the neural network for a limited number of iterations.
\item[Step 2] Estimate the mean PDE residual $\mathcal{E}_r$ in \cref{eq:rat} by Monte Carlo integration, i.e., by the average of values at a set of randomly sampled locations $\mathcal{S} = \{\mathbf{x}_1, \mathbf{x}_2, \dots, \mathbf{x}_{|\mathcal{S}|}\}$:
\begin{equation*}
    \mathcal{E}_r \approx \frac{1}{|\mathcal{S}|} \sum_{\mathbf{x} \in \mathcal{S}} \left\Vert f\left( \mathbf{x}; \frac{\partial \hat{u}}{\partial x_1},  \dots, \frac{\partial \hat{u}}{\partial x_d}; \frac{\partial^2 \hat{u}}{\partial x_1\partial x_1}, \dots, \frac{\partial^2 \hat{u}}{\partial x_1\partial x_d};\dots;\bm{\lambda} \right) \right\Vert.
\end{equation*}
\item[Step 3] Stop if $\mathcal{E}_r < \mathcal{E}_0$. Otherwise, add $m$ new points with the largest residuals in $\mathcal{S}$ to $\mathcal{T}$, and go to Step 2.
\end{enumerate}
\end{algorithm}

\section{DeepXDE usage and customization}
\label{sec:deepxde}

In this section, we introduce the usage of DeepXDE and how to customize DeepXDE to meet new problem requirements.  

\subsection{Usage}

\begin{figure}[htbp]
    \centering
    \includegraphics{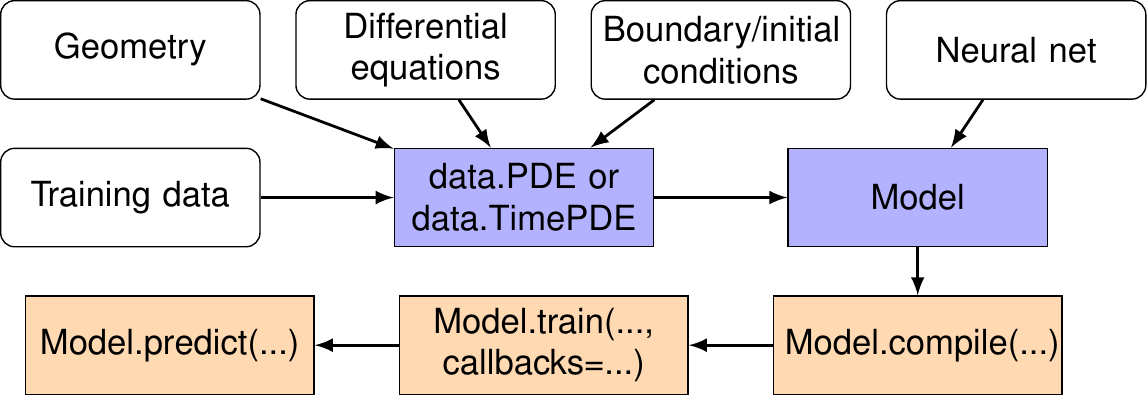}
    \caption{Flowchart of DeepXDE corresponding to Procedure~\ref{alg:deepxde}. The white boxes define the PDE problem and the training hyperparameters. The blue boxes combine the PDE problem and training hyperparameters in the white boxes. The orange boxes are the three steps (from right to left) to solve the PDE.}
    \label{fig:DeepXDE}
\end{figure}

Compared to traditional numerical methods, the code written with DeepXDE is much shorter and more comprehensive, resembling closely the mathematical formulation. Solving differential equations in DeepXDE is no more than specifying the problem using the build-in modules, including computational domain (geometry and time), PDE equations, boundary/initial conditions, constraints, training data, neural network architecture, and training hyperparameters. The workflow is shown in Procedure~\ref{alg:deepxde} and \cref{fig:DeepXDE}.

\begin{algorithm}[htbp]
\caption{Usage of DeepXDE for solving differential equations.}
\label{alg:deepxde}
\begin{enumerate}
\item[Step 1] Specify the computational domain using the \textbf{\texttt{geometry}} module.
\item[Step 2] Specify the PDE using the grammar of \textbf{\texttt{TensorFlow}}.
\item[Step 3] Specify the boundary and initial conditions.
\item[Step 4] Combine the geometry, PDE and boundary/initial conditions together into \textbf{\texttt{data.PDE}} or \textbf{\texttt{data.TimePDE}} for time-independent problems or  for time-dependent problems, respectively. To specify training data, we can either set the specific point locations, or only set the number of points and then DeepXDE will sample the required number of points on a grid or randomly.
\item[Step 5] Construct a neural network using the \textbf{\texttt{maps}} module.
\item[Step 6] Define a \textbf{\texttt{Model}} by combining the PDE problem in Step 4 and the neural net in Step 5.
\item[Step 7] Call \textbf{\texttt{Model.compile}} to set the optimization hyperparameters, such as optimizer and learning rate. The weights in \cref{eq:loss} can be set here by \textbf{\texttt{loss\_weights}}.
\item[Step 8] Call \textbf{\texttt{Model.train}} to train the network from random initialization or a pre-trained model using the argument \textbf{\texttt{model\_restore\_path}}. It is extremely flexible to monitor and modify the training behavior using \textbf{\texttt{callbacks}}.
\item[Step 9] Call \textbf{\texttt{Model.predict}} to predict the PDE solution at different locations.
\end{enumerate}
\end{algorithm}

\begin{figure}[htbp]
    \centering
    \includegraphics{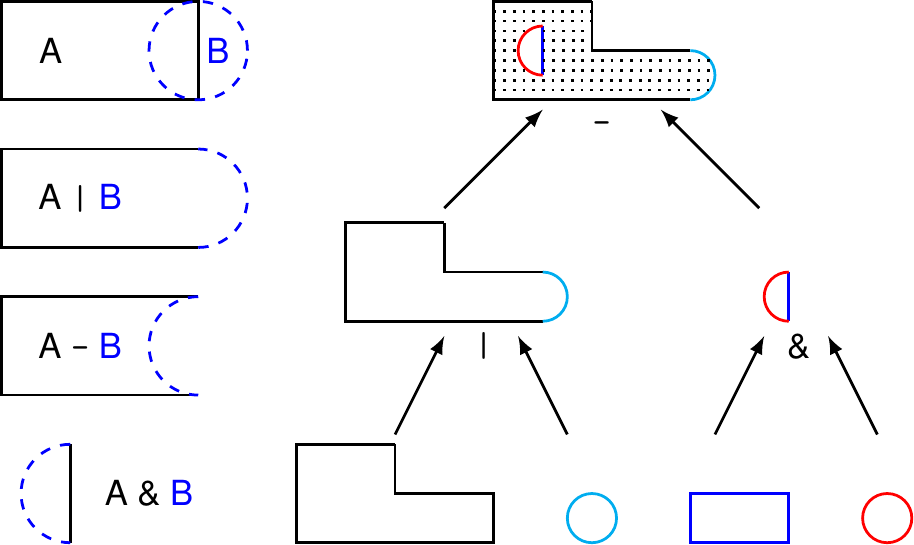}
    \caption{Examples of constructive solid geometry (CSG) in 2D. (\textbf{left}) A and B represent the rectangle and circle, respectively. The union $A|B$, difference $A-B$, and intersection $A\& B$ are constructed from A and B. (\textbf{right}) A complex geometry (top) is constructed from a polygon, a rectangle and two circles (bottom) through the union, difference, and intersection operations. This capability is included in the module \texttt{geometry} of DeepXDE.}
    \label{fig:csg}
\end{figure}

In DeepXDE, the built-in primitive geometries include \textbf{\texttt{interval}}, \textbf{\texttt{triangle}}, \textbf{\texttt{rectangle}}, \textbf{\texttt{polygon}}, \textbf{\texttt{disk}}, \textbf{\texttt{cuboid}} and \textbf{\texttt{sphere}}. Other geometries can be constructed from these primitive geometries using three boolean operations: \textbf{\texttt{union (|)}}, \textbf{\texttt{difference (-)}} and \textbf{\texttt{intersection (\&)}}. This technique is called constructive solid geometry (CSG), see \cref{fig:csg} for examples. CSG supports both two-dimensional and three-dimensional geometries.

DeepXDE supports four standard boundary conditions, including Dirichlet (\textbf{\texttt{DirichletBC}}), Neumann (\textbf{\texttt{NeumannBC}}), Robin (\textbf{\texttt{RobinBC}}), and periodic (\textbf{\texttt{PeriodicBC}}), and a more general BC can be defined using \textbf{\texttt{OperatorBC}}. The initial condition can be defined using \textbf{\texttt{IC}}. There are two types of neural networks available in DeepXDE: feed-forward neural network (\textbf{\texttt{maps.FNN}}) and residual neural network (\textbf{\texttt{maps.ResNet}}). It is also convenient to choose different training hyperparameters, such as loss types, metrics, optimizers, learning rate schedules, initializations and regularizations.

In addition to solving differential equations, DeepXDE can also be used to approximate functions from multi-fidelity data \cite{meng2019composite}, and learn nonlinear operators \cite{lu2019deeponet}.

\subsection{Customizability}

All the components of DeepXDE are loosely coupled, and thus DeepXDE is well-structured and highly configurable. In this subsection, we discuss how to customize DeepXDE to address new problem requirements, e.g., new geometry or network architecture.

\subsubsection{Geometry}

As we introduced above, DeepXDE has already supported 7 basic geometries and the CSG technique. However, it is still possible that the user needs a new geometry, which cannot be constructed in DeepXDE. In this situation, a new geometry can be defined as shown in Procedure~\ref{code:geometry}. Currently DeepXDE does not support accurately descriptions of complex curvilinear boundaries; however, a future extension could be incorporation of the non-uniform rational basis spline (NURBS) \cite{hughes2005isogeometric} for such representations.

\begin{algorithm}[htbp]
\caption{Customization of the new geometry module \textbf{\texttt{MyGeometry}}. The class methods should only be implemented as needed.}\label{code:geometry}
{\footnotesize
\begin{minted}[linenos]{Python}
class MyGeometry(Geometry):
    def inside(self, x):
        """Check if x is inside the geometry."""
    def on_boundary(self, x):
        """Check if x is on the geometry boundary."""
    def boundary_normal(self, x):
        """Compute the unit normal at x for Neumann or Robin boundary conditions."""
    def periodic_point(self, x, component):
        """Compute the periodic image of x for periodic boundary condition."""
    def uniform_points(self, n, boundary=True):
        """Compute the equispaced point locations in the geometry."""
    def random_points(self, n, random="pseudo"):
        """Compute the random point locations in the geometry."""
    def uniform_boundary_points(self, n):
        """Compute the equispaced point locations on the boundary."""
    def random_boundary_points(self, n, random="pseudo"):
        """Compute the random point locations on the boundary."""
\end{minted}
}
\end{algorithm}

\subsubsection{Neural networks}

DeepXDE currently supports two neural networks: feed-forward neural network (\textbf{\texttt{maps.FNN}}) and residual neural network (\textbf{\texttt{maps.ResNet}}). A new network can be added as shown in Procedure~\ref{code:nn}.

\begin{algorithm}[htbp]
\caption{Customization of the neural network \textbf{\texttt{MyNet}}.}\label{code:nn}
{\footnotesize
\begin{minted}[linenos]{Python}
class MyNet(Map):
    @property
    def inputs(self):
        """Return the net inputs."""
    @property
    def outputs(self):
        """Return the net outputs."""
    @property
    def targets(self):
        """Return the targets of the net outputs."""
    def build(self):
        """Construct the network."""
\end{minted}
}
\end{algorithm}

\subsubsection{Callbacks}

It is usually a good strategy to monitor the training process of the neural network, and then make modifications in real time, e.g., change the learning rate. In DeepXDE, this can be implemented by adding a callback function, and here we only list a few commonly used ones already implemented in DeepXDE:
\begin{itemize}
    \item \textbf{\texttt{ModelCheckpoint}}, which saves the model after certain epochs or when a better model is found.
    \item \textbf{\texttt{OperatorPredictor}}, which calculates the values of the operator applying on the outputs.
    \item \textbf{\texttt{FirstDerivative}}, which calculates the first derivative of the outpus with respect to the inputs. This is a special case of \textbf{\texttt{OperatorPredictor}} with the operator being the first derivative.
    \item \textbf{\texttt{MovieDumper}}, which dumps the movie of the function during the training progress, and/or the movie of the spectrum of its Fourier transform.
\end{itemize}
It is very convenient to add other \textbf{\texttt{callback}} functions, which will be called at different stages of the training process, see Procedure~\ref{code:callback}.

\begin{algorithm}[htbp]
\caption{Customization of the callback \textbf{\texttt{MyCallback}}. Here, we only show how to add functions to be called at the beginning/end of every epoch. Similarly, we can call functions at the other training stages, such as at the beginning of training.}\label{code:callback}
{\footnotesize
\begin{minted}[linenos]{Python}
class MyCallback(Callback):
    def on_epoch_begin(self):
        """Called at the beginning of every epoch."""
    def on_epoch_end(self):
        """Called at the end of every epoch."""
\end{minted}
}
\end{algorithm}

\section{Demonstration examples}
\label{sec:example}

In this section, we use PINNs and DeepXDE to solve different problems. In all examples, we use the $\tanh$ as the activation function, and the other hyperparameters are listed in \cref{tab:hyperparameter}. The weights $w_f$, $w_b$ and $w_i$ in the loss function are set as 1. The codes of all examples are published in GitHub.

\begin{table}[htbp]
{\footnotesize
  \caption{Hyperparameters used for the following 5 examples. ``Adam, L-BFGS'' represents that we first use Adam for a certain number of iterations, and then switch to L-BFGS. The optimizer L-BFGS does not require learning rate, and the neural network is trained until convergence, so the number of iterations is also ignored for L-BFGS.}
  \label{tab:hyperparameter}
\begin{center}
  \begin{tabular}{c|ccccc} \hline
   Example & NN Depth & NN Width & Optimizer & Learning rate & \# Iterations \\ \hline
    1 & 4 & 50 & Adam, L-BFGS & 0.001 & 50000\\
    2 & 3 & 20 & Adam, L-BFGS & 0.001 & 15000 \\
    3 & 3 & 40 & Adam & 0.001 & 60000 \\
    4 & 3 & 20 & Adam & 0.001 & 80000 \\
    5 & 4 & 20 & L-BFGS & - & - \\ \hline
  \end{tabular}
\end{center}
}
\end{table}

\subsection{Poisson equation over an L-shaped domain}
\begin{figure}[htbp]
    \centering
    \includegraphics[width=\textwidth]{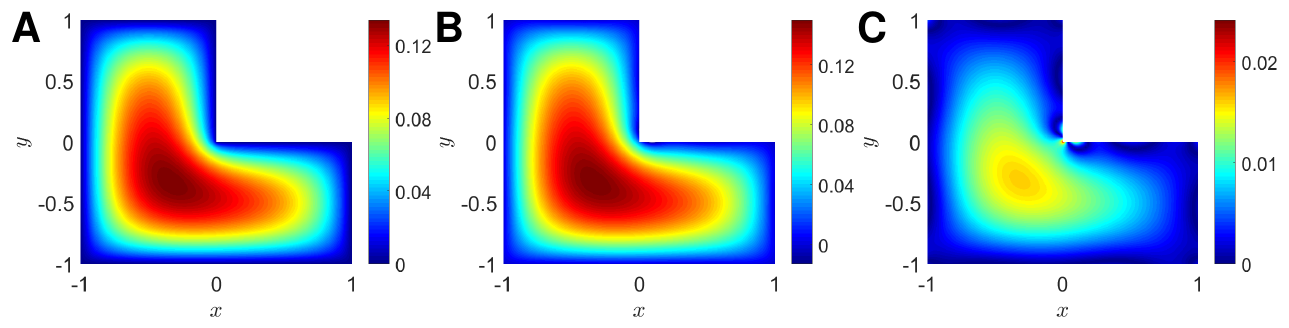}
    \caption{Example 4.1. Comparison of the PINN solution with the solution obtained by using spectral element method (SEM). (\textbf{A}) the SEM solution $u_{SEM}$, (\textbf{B}) the PINN solution $u_{NN}$, (\textbf{C}) the absolute error $|u_{SEM} - u_{NN}|$.}
    \label{fig:Poisson:Lshape}
\end{figure}

Consider the following two-dimensional Poisson equation over an L-shaped domain $\Omega = [-1,1]^2\setminus [0,1]^2$:
\begin{align*}
    -\Delta u(x,y) &= 1,\quad (x,y)\in \Omega, \\
    u(x,y) &= 0, \quad  (x,y)\in \partial \Omega.
\end{align*}
We choose 1200 and 120 random points drawn from a uniform distribution as $\mathcal{T}_f$ and $\mathcal{T}_b$, respectively. The PINN solution is given in \cref{fig:Poisson:Lshape}B. For comparison, we also present the numerical solution obtained by using the  spectral element method (SEM)~\cite{Karbook} (\cref{fig:Poisson:Lshape}A). The result of the absolute error is shown in \cref{fig:Poisson:Lshape}C.

\subsection{RAR for Burgers equation}
\begin{figure}[htbp]
    \centering
    \includegraphics{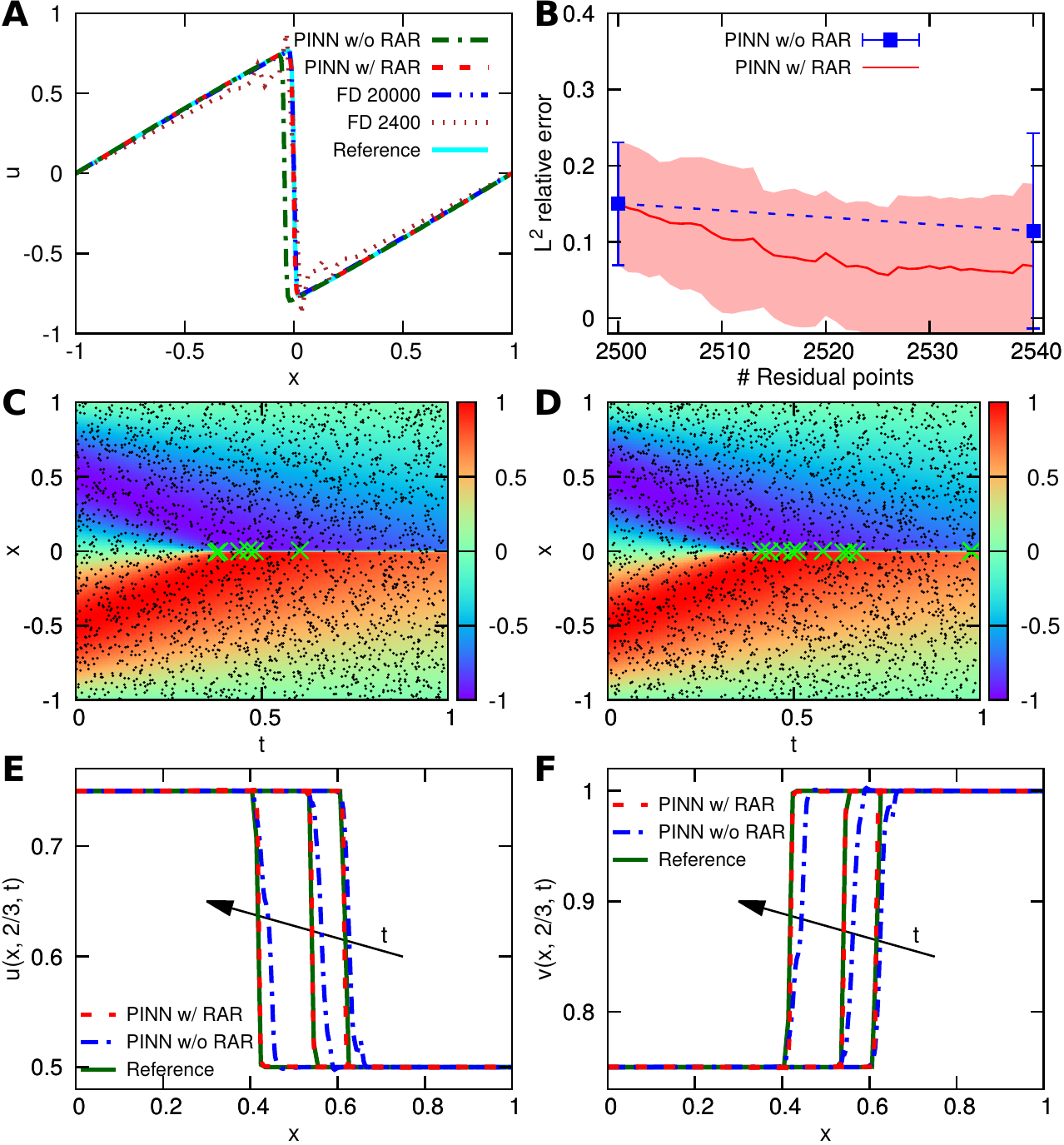}
    \caption{Example 4.2. Comparisons of the PINN solutions of (A, B, C and D) 1D and (E and F) 2D Burgers equations with and without RAR. (\textbf{A}) The cyan, green and red lines represent the reference solution of $u$ from \cite{raissi2019physics}, PINN solution without RAR, PINN solution with RAR at $t = 0.9$, respectively. For the finite difference (FD) method, $200\times 100 = 20000$ spatial-temporal grid points are used to achieve a good solution (blue line). If only $60\times 40 = 2400$ points are used, the FD solution has large oscillations around the discontinuity (brown line). (\textbf{B}) $L^2$ relative error versus the number of residual points. The red solid line and shaded region correspond to the mean and one-standard-deviation band for the $L^2$ relative error of PINN with RAR, respectively. The blue dashed line is the mean and one-standard-deviation for the error of PINN using 2540 random residual points. The mean and standard deviation are obtained from 10 runs with random initial residual points. (\textbf{C} and \textbf{D}) Two representative examples for the residual points added via RAR. Black dots: initial residual points; green cross: added residual points; 6 and 11 residual points are added in C and D, respectively. (\textbf{E} and \textbf{F}) Comparison of the velocity profiles at $y = \frac{2}{3}$ from the PINNs with and without RAR for the 2D Burgers equation. The profiles at three different times ($t = 0.2$, 0.5 and 1) are presented with $t$ increasing along the direction of the arrow.}
    \label{fig:burgers}
\end{figure}

We first consider the 1D Burgers equation: 
\begin{gather*}
    \frac{\partial u}{\partial t} + u \frac{\partial u}{\partial x} = \nu \frac{\partial^2 u}{\partial x^2}, \quad x\in[-1, 1], ~ t \in[0, 1], \\
    u(x, 0) = -\sin(\pi x), \quad  u(-1, t) = u(1, t) = 0.
\end{gather*}
Let $\nu = 0.01/\pi$. Initially, we randomly select 2500 points
(spatio-temporal domain) as the residual points, and then 40 more residual points are added adaptively via RAR developed in \cref{sec:rar} with $m = 1$ and $\mathcal{E}_0 = 0.005$. We compare the PINN solution with RAR and the PINN solution based on 2540 randomly selected training data (\cref{fig:burgers} A and B), and demonstrate that PINN with RAR can capture the discontinuity much better. For a comparison, the finite difference solutions using central difference scheme for space discretization and forward Euler scheme for time discretization for the Burgers equation in the conservative form are also shown in \cref{fig:burgers}A. Here, we also present two examples for the residual points added via the RAR method. As shown in \cref{fig:burgers} C and D, the added points (green crosses) are quite close to the sharp interface, which indicates the effectiveness of RAR.

We further solve the following two-dimensional Burgers equation using the RAR:
\begin{gather*}
    \partial_t u + u \partial_x u + v \partial_y u = \frac{1}{\mbox{Re}} (\partial^2_x u + \partial^2_y u),\\
    \partial_t v + u \partial_x v + v \partial_y v = \frac{1}{\mbox{Re}} (\partial^2_x v + \partial^2_y v),\\
    x, y \in [0, 1], ~ \mbox{and} ~t \in [0, 1],
\end{gather*}
where $u$ and $v$ are the velocities along the $x$ and $y$ directions, respectively. In addition, Re is a non-dimensional number (Reynolds number) defined as Re $= UL/\nu$, in which $U$ and $L$ are respectively the characteristic velocity and length, and $\nu$ is the kinematic viscosity of fluid. The exact solution can be obtained \cite{baeza2006adaptive} as
\begin{gather*}
    u(x, y, t) = \frac{3}{4} - \frac{1}{4\left[ 1 + \exp((-4x + 4y - t)\mbox{Re}/32) \right]}, \\
    v(x, y, t) = \frac{3}{4} + \frac{1}{4\left[ 1 + \exp((-4x + 4y - t)\mbox{Re}/32) \right]},
\end{gather*}
using the Dirichlet boundary conditions on all boundaries. In the present study, Re is set to be 5000, which is quite challenging due to the fact that the high Reynolds number leads to steep gradient in the solution. Initially, 200 residual points are randomly sampled in the spatio-temporal domain, and 5000 and 1000 random points are used for each initial and boundary conditions, respectively. We only add 10 extra residual points using RAR, and the total number of the residual points is 210 after convergence. For comparison, we also test the case without the RAR using 210 randomly sampled residual points. The results are displayed in \cref{fig:burgers} E and F, demonstrating the effectiveness of RAR.

\subsection{Inverse problem for the Lorenz system}
\begin{figure}[htbp]
    \centering
    \includegraphics{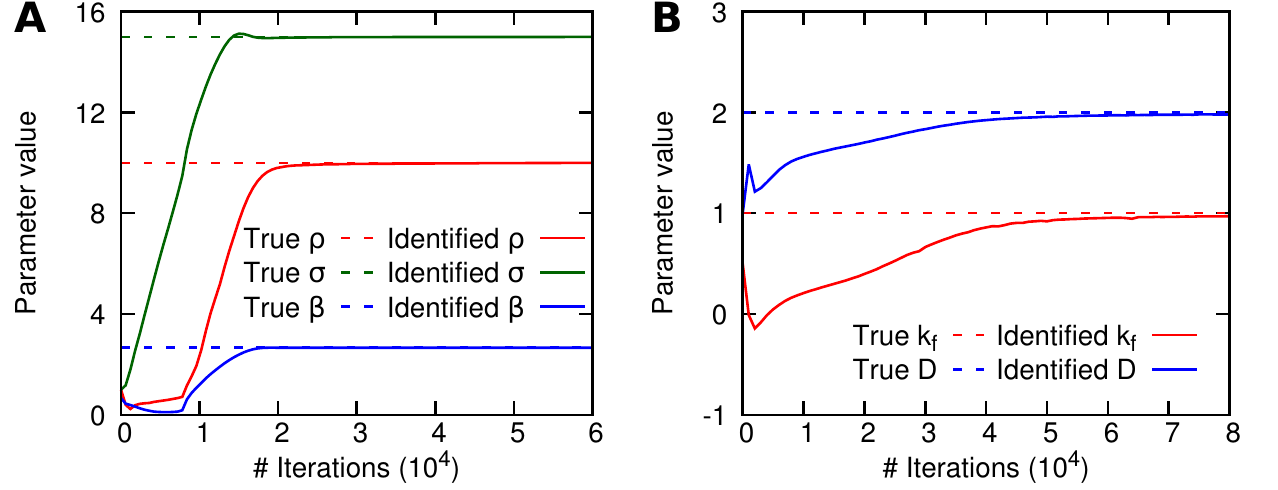}
    \caption{Examples 4.3 and 4.4. The identified values of (\textbf{A}) the Lorenz system and (\textbf{B}) diffusion-reaction system converge to the true values during the training process. The parameter values are scaled for plotting.}
    \label{fig:inverse}
\end{figure}

Consider the parameter identification problem of the following Lorenz system 
\begin{equation*}
    \frac{dx}{dt} = \rho(y-x), \quad
    \frac{dy}{dt} = x(\sigma-z)-y, \quad
    \frac{dz}{dt} = xy-\beta z,
\end{equation*}
with the initial condition $(x(0), y(0), z(0)) = (-8,7,27)$, where $\rho$, $\sigma$ and $\beta$ are the three parameters to be identified from the observations at certain times. The observations are produced by solving the above system to $t = 3$ using Runge-Kutta (4,5) with the underlying true parameters $(\rho, \sigma, \beta) = (10, 15, 8/3)$. We choose 400 uniformly distributed random points and 25 equispaced points as the residual points $\mathcal{T}_f$ and $\mathcal{T}_i$, respectively. The evolution trajectories of $\rho$, $\sigma$ and $\beta$ are presented in \cref{fig:inverse}A, with the final identified values of $(\rho, \sigma, \beta) = (10.002, 14.999, 2.668)$.

\subsection{Inverse problem for diffusion-reaction systems} 

A diffusion-reaction system in porous media for the solute concentrations $C_A$, $C_B$ and $C_C$ ($A + 2B \rightarrow C$) is described by
\begin{gather*}
\frac{\partial C_A}{\partial t} = D \frac{\partial^2 C_A}{\partial x^2} - k_f C_A C^2_B, \quad
\frac{\partial C_B}{\partial t} = D \frac{\partial^2 C_B}{\partial x^2} - 2 k_f C_A C^2_B, \quad x \in [0, 1], t \in[0, 10], \\
C_A(x, 0) = C_B(x, 0) = e^{-20x}, \quad C_A(0, t) = C_B(0, t) = 1, \quad C_A(1, t) = C_B(1, t) = 0,
\end{gather*}
where $D = 2 \times 10^{-3}$ is the effective diffusion coefficient, and $k_f = 0.1$ is the effective reaction rate. Because $D$ and $k_f$ depend on the pore structure and are difficult to measure directly, we estimate $D$ and $k_f$ based on 40000 observations of the concentrations $C_A$ and $C_B$ in the spatio-temporal domain. The identified $D$ ($1.98 \times 10^{-3}$) and $k_f$ (0.0971) are displayed in \cref{fig:inverse}B, which agree well with their true values.

\subsection{Volterra IDE}

\begin{figure}[htbp]
    \centering
    \includegraphics{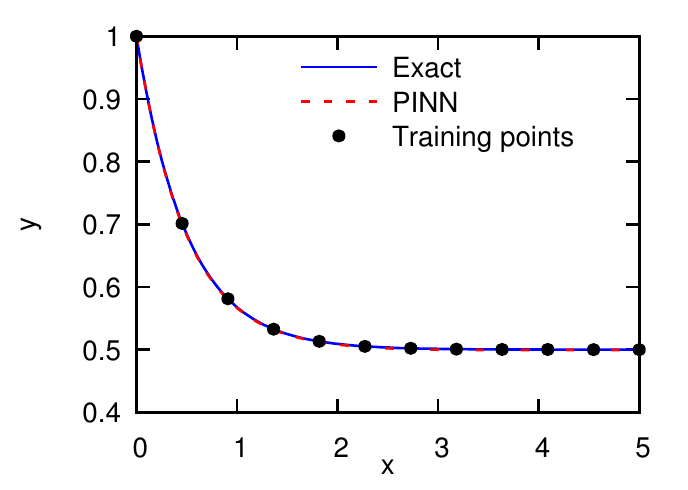}
    \caption{Example 4.5. The PINN algorithm for solving Volterra IDE. The blue solid line is the exact solution, and the red dash line is the numerical solution from PINN. 12 equispaced residual points (black dots) are used.}
    \label{fig:volterra}
\end{figure}

Here, we consider the first-order integro-differential equation of the Volterra type in the domain $[0, 5]$:
$$\frac{dy}{dx} + y(x) = \int_0^x e^{t-x}y(t)dt, \quad y(0) = 1,$$
with the exact solution $y(x) = e^{-x}\cosh x.$ We solve this IDE using the method in \cref{sec:ide}, and approximate the integral using Gaussian-Legendre quadrature of degree 20. The $L^2$ relative error is $2 \times 10^{-3}$, and the solution is shown in \cref{fig:volterra}.

\section{Concluding Remarks}
\label{sec:conc}

In this paper, we present the algorithm, approximation theory, and error analysis of the physics-informed neural networks (PINNs) for solving different types of partial differential equations (PDEs). Compared to the traditional numerical methods, PINNs employ automatic differentiation to handle differential operators, and thus they are mesh-free. Unlike numerical differentiation, automatic differentiation does not differentiate the data and hence it can tolerate noisy data for training. We also discuss how to extend PINNs to solve other types of differential equations, such as integro-differential equations, and also how to solve inverse problems. In addition, we propose a residual-based adaptive refinement (RAR) method to improve the distribution of residual points during the training process, and thus increase the training efficiency.

To benefit both the education and the computational science communities, we have developed the Python library DeepXDE, an implementation of PINNs. By introducing the usage of DeepXDE, we show that DeepXDE enables user codes to be compact and follow closely the mathematical formulation. We also demonstrate how to customize DeepXDE to meet new problem requirements. Our numerical examples for forward and inverse problems verify the effectiveness of PINNs and the capability of DeepXDE. Scientific machine learning is emerging as a new and potentially powerful alternative to classical scientific computing, so we hope that libraries such as DeepXDE will accelerate this development and will make it accessible to the classroom but also to other researchers who may find the need to adopt PINN-like methods in their research, which can be very effective especially for inverse problems. 

Despite the aforementioned advantages, PINNs still have some limitations. For forward problems, PINNs are currently slower than finite elements but this can be alleviated via offline training \cite{zhu2019physics,winovich2019convpde}. For long time integration, one can also use time-parallel methods to simultaneously compute on multiple GPUs for shorter time domains. Another limitation is the search for effective neural network architectures, which is currently done empirically by users; however, emerging meta-learning techniques can be used to automate this search, see \cite{zoph2016neural,finn2017model}. Moreover, while here we enforce the strong form of PDEs, which is easy to be implemented by automatic differentiation, alternative weak/variational forms may also be effective, although they require the use of quadrature grids. Many other extensions for multi-physics and multi-scale problems are possible across different scientific disciplines by creatively designing the loss function and introducing suitable solution spaces. For instance, in the five examples we present here, we only assume data on scattered points, however, in geophysics or biomedicine we may have mixed data in the form of images and point measurements. In this case, we can design a composite neural network consisting of one convolutional neural network and one PINN sharing the same set of parameters, and minimize the total loss which could be a weighted summation of multiple losses from each neural network.

\section*{Acknowledgments}

This work is supported by the DOE PhILMs project (No. de-sc0019453), the AFOSR grant FA9550-17-1-0013, and the DARPA-AIRA grant HR00111990025.

\bibliographystyle{siamplain}
\bibliography{main}
\end{document}

%% file: main_shared.tex
\usepackage{lipsum}
\usepackage{amsfonts}
\usepackage{graphicx}
\usepackage{epstopdf}
\usepackage{algorithmic}
\ifpdf
  \DeclareGraphicsExtensions{.eps,.pdf,.png,.jpg}
\else
  \DeclareGraphicsExtensions{.eps}
\fi

\newsiamremark{remark}{Remark}
\newsiamremark{hypothesis}{Hypothesis}
\crefname{hypothesis}{Hypothesis}{Hypotheses}
\newsiamthm{claim}{Claim}

\headers{DeepXDE}{L. Lu, X. Meng, Z. Mao, and G. E. Karniadakis}

\title{DeepXDE: A deep learning library for solving differential equations}

\author{Lu Lu\thanks{Division of Applied Mathematics, Brown University, Providence, RI (\email{george\_karniadakis@brown.edu}).}
\and Xuhui Meng\footnotemark[1]
\and Zhiping Mao\footnotemark[1]
\and George Em Karniadakis$^*$\thanks{Pacific Northwest National Laboratory, Richland, WA}}

\usepackage{amsopn}
